\documentclass[12pt]{article}
\usepackage{amsmath, amsthm, amssymb}
\usepackage[USenglish]{babel}
\usepackage{amsmath}
\usepackage{mathtools}
\usepackage{graphicx}
\usepackage{booktabs}
\usepackage{verbatim}
\usepackage{placeins}
\usepackage{adjustbox}
\usepackage{setspace}
\usepackage{url}
\usepackage{tabularx}
\usepackage{pdflscape}
\usepackage{booktabs,caption}
\usepackage[flushleft]{threeparttable}
\usepackage[normalem]{ulem}
\usepackage{a4wide}
\def\sym#1{\ifmmode^{#1}\else\(^{#1}\)\fi}
\usepackage{caption}
\useunder{\uline}{\ul}{}

\title{\large Examining the Presence of Gender Bias in Customer Reviews Using Word Embedding} 

\date{}

\usepackage[top=2.5cm, bottom=1.5cm, left=2.5cm, right=2cm]{geometry}

\author{Arul Mishra, Himanshu Mishra, Shelly Rathee\footnote{Authors are at David Eccles School of Business, Department of Marketing, University of Utah, Salt Lake City, Utah - 84112}}

\begin{document}

\maketitle

\begin{abstract}
Humans have entered the age of algorithms. Each minute, algorithms shape countless preferences from suggesting a product to a potential life partner. In the marketplace, algorithms are trained to learn consumer preferences from customer reviews because user-generated reviews are considered the voice of customers and a valuable source of information to firms. Insights, minded from reviews, play an indispensable role in several business activities ranging from product recommendations, targeted advertising, promotions, segmentation, etc. In this research, we question whether reviews might hold stereotypic gender bias that algorithms learn and propagate. Utilizing data from millions of observations and a word embedding approach, GloVe, we show that algorithms designed to learn from human language output, also learn gender bias. We also examine why such biases occur: whether the bias is caused because of a negative bias against females or a positive bias for males. We examine the impact of gender bias in reviews on choice and conclude with policy implications for female consumers, especially when they are unaware of the bias, and the ethical implications for the firms.         
\end{abstract}



\newpage
Firms extensively use natural language processing algorithms to gather insights. Insights mined from reviews play an indispensable role in several business activities ranging from product recommendation, targeted advertising, promotions, segmentation, etc. For instance, algorithmic recommendations have been ubiquitous. They are used for guiding consumer decisions ranging from what brand to buy, places to visit, to even job offerings or a potential life partner. While there is plethora of research on the importance of reviews, there is no work that documents the presence of a bias in reviews, especially gender bias. In this research, we apply a recent word-embedding algorithm, GloVe (Global Vector), to more than 11 million reviews from Amazon and Yelp. Our findings indicate the presence of gender bias in reviews. Along with demonstrating the presence of gender bias, we examine whether gender bias exists because of a bias against women or a bias for men. Interventions can be successful only when the nature of the gender bias is known. Overall, we find that women are more associated with negative (e.g. fickle, impulsive, homebound) rather than positive attributes (e.g. loyal, sensible, industrious). However, men are more associated with positive rather than negative attributes. In order to address the question of whether gender bias in reviews have an influence on consumer decisions, we focus on a specific marketing domain, product recommendations, and demonstrate that consumer choice especially that of female consumers does get impacted. Firms routinely gather insights from customer reviews assuming them to be the voice-of-the-customer but if these reviews hold gender bias, then any decision made using those reviews has the power to harm customers. Biased input can result in biased recommendations, messages, or insensitive portrayals, and could result in learning and using consumer vulnerability against female consumers.

Insights mined from reviews play an indispensable role in several business activities ranging from product recommendation, targeted advertising, promotions, segmentation etc. For instance, algorithmic recommendations have become ubiquitous. They are used for guiding consumer decisions ranging from which brand to buy, books to read, clothes to wear, foods to eat, places to visit, to even job offerings or a potential life partner (Jannach et al. 2010; Schrage 2014). Their ubiquity comes from the fact that consumer preferences can be nudged and shaped by these recommendations (Häubl and Murray 2006). While some recommendations rely on prior purchase behavior, the bulk of recommendations are made for products that a consumer has never purchased before. Consider travel planning - for 63\% of consumers who make travel plans after searching the Internet, booking a hotel follows a standard pattern online (Rheem 2012). They search for the destination based on their vacation dates and see numerous booking sites that Google shows them. The travel site has very little information about the consumers’ prior travel history or preferences except for some demographic information like gender and prior browsing history that Google shares with the travel site. In such situations, the recommendations by the travel site are invariably based on how similar a consumer might be to available consumer profiles, based on data shared by Google. How are these profiles created? Algorithms create these profiles by analyzing millions of reviews generated by other consumers. Analysis of prior reviews enables a site to make recommendations to consumers with no prior information about their preferences (Levi et al. 2012).

This marketplace example highlights the fundamental role textual analysis of consumer reviews holds in creating profiles of potential consumers (Adomavicius and Tuzhilin 2015; Hariri et al. 2011). Customer generated reviews are critical for their role in providing customized recommendations to consumers (Dong et al. 2015; Levi et al. 2012; Sun et al. 2009). Since, reviews reflect consumers’ post-purchase consumption experience, they play an indispensable role in several business activities ranging from targeted advertising, promotions, segmentation etc. (Lee 2009; Netzer et al. 2012; Sun et al. 2009).
As existing marketing literature documents, it makes logical sense for businesses to use reviews in different business activities because consumers themselves rely on other consumers’ reviews to make decisions (Aciar et al. 2007; Jakob et al. 2009). Reviews are considered the “voice of customers” by firms (Griffin and Hauser 1993) because they inform firms about consumer opinions, sentiments, needs, (Humphreys and Wang 2017) and can be used as a voluntary source of market research data (Lee and Bradlow 2011). Reviews are so important that companies such as Amazon and Yelp constantly moderate the content and act promptly if there is any tampering with the reviews\footnote{https://www.amazon.com/gp/help/customer/display.html?nodeId=201929730}.

The use of reviews in any marketplace decision e.g. to make recommendation to consumers, segmentation etc. relies on the premise that reviews provide a window into consumers’ mind that can help the firm better satisfy consumer needs by customizing the offerings and hence maximize profits. While there are hundreds of academic articles written on the importance of reviews, to the best of our knowledge there is virtually no work that documents the presence of any bias in reviews especially any unfair difference in the way women and men are portrayed.

Specifically, in this research we demonstrate that user-generated reviews hold stereotypic gender bias. Even on websites that monitor reviews for offensive content, we find the existence of gender bias. Our analysis of more than 11 million reviews from Amazon and Yelp, using a recent word-embedding algorithm, GloVe (Global Vector), indicates that the profile of women is more associated with negative (e.g. fickle, impulsive, lazy, homebound) rather than positive attributes (e.g. loyal, sensible, industrious, career oriented).

We not only investigate whether customer reviews may hold stereotypic bias, but also examine the nature of the bias that might affect firm decisions. Interventions can be successful only if we know the nature of the bias. Past research has shown that one form of bias that exists against women is because they are perceived to be more associated with family and less so with career (Greenwald, McGhee, and Schwartz 1998). Therefore in our research, we examine whether gender bias in customer reviews exists because of a bias against women (considering them less suitable for a career but more suitable for home) or a bias for men (considering them more suitable for a career but less suitable for home). 
Reviews are consequential for firms because they are used for many business activities ranging from product design, promotions, segmentation, targeting, recommendations etc. To examine what harm can be caused if reviews hold gender bias, we further focus on a specific use of reviews: product recommendations. Mimicking how gender bias could influence recommendations, we first identify products that are associated with positive and negative labels (e.g. impulsive/vain versus sensible/determined) and then recommend these labeled products to both male and female consumers. If consumers were more likely to choose the recommended product, it would demonstrate the harm caused by the gender bias in reviews, which are used to create the recommendations.

\section*{\large Ethical Consequences}

The presence of bias in reviews may seem unsurprisingly consistent with the portrayal of women in media and advertising. For decades, research has documented that media portrays women as the more dependent gender (Ford et al. 1998). Since reviews are not written in a vacuum, they reflect the cultural milieu. However, one can argue that since product reviews are considered the voice-of-the-customer, it doesn’t matter whether they hold gender bias. Even with biases, reviews do provide insights into the minds of the consumer. The problem with such an argument is that if a firm uses reviews, containing gender bias to design products and promotions, recommend products to consumers, determine how to portray individuals in a commercial, segment consumers etc. then it could lead to biased recommendations, messages, and insensitive portrayals. Business ethics relate to ensuring that firms know the consequences of their decision and individuals are not harmed by the decisions of a firm. For instance, if recommendations are designed based on biased reviews, the consequence can be unethical: using consumer vulnerability (e.g. impulsivity) to persuade consumers towards a certain choice against them. If the reviews contain bias, then the recommendations are biased and unethical, and could result in legal ramifications against the firm. Biased reviews, which are used to make product recommendations and to make any business decisions, can acquire a troubling distinction when we consider the following two aspects unique to algorithms, influencing society.

First, algorithms learn, amplify, and propagate. An example is Google translate: entering a gender-neutral sentence in Hindi such as “Vah ek doktar hai” (“That is a doctor”) gets translated in English as “He is a doctor.” Similarly, “Vah ek nurse hai” (“That is a nurse”) gets translated in English as “She is a nurse.” While, it is true that there are more male doctors than female doctors due to rates of burnout (Poorman 2018), the algorithm is interpreting it as a stronger association of males with doctors and females with nurses. Thus, the stereotype that doctors are male and nurses are female is perpetuated. Another example is when YouTube’s autocomplete algorithm, which relies on prior textual searches by other users to predict what is being typed, began providing pedophilic suggestions (Graham 2017). The impact of propagating false stereotypes is emerging in job search algorithms as well. Incorrectly learning a social disparity that women are overrepresented in lower paying jobs as women’s preference for lower paying jobs, the algorithms recommend lower paying jobs to women than to men (Bolukbasi et al. 2016). This leads to an unavoidable feedback loop that would only lead to higher female representation in lower paying jobs. Recent research underlines such a disturbing trend when it finds that online posts on a question-answer site receive a lower evaluation if associated with a female rather than male user name (Bohren, Imas, and Rosenberg 2018).

Second is the algorithmic feature of trust. Algorithms are considered neutral experts, whose suggestions people trust, making their recommendations consequential (Hoffman and Novak 2017; Robinette et al. 2016). People tend to perceive algorithms as having vast resources to synthesize information and consider them experts on many topics. Moreover, people consider algorithms an objective source that help humans in myriad tasks ranging from border patrol to product selection (Yang et al. 2017). People blindly follow algorithm suggestions even when algorithms actually are making mistakes that could endanger their lives (Robinette et al. 2016). The overall belief is that because algorithms are machines, they are likely to be more objective than humans. Since, firms use information from customer reviews to make recommendations and consumers trust algorithmic recommendations, the problem becomes exacerbated. 

\vspace{6mm}
\section*{\large Research Overview}

Procedurally, we first discuss literature that demonstrates that gender bias, especially implicit, can influence behaviors that can be biased against a group. We discuss this biased behavior, in the marketplace and business world. Second, since language captures current social and cultural feelings, we describe how implicit bias can insidiously creep into consumer reviews. Third, using a recent word-embedding method, GloVe (Global Vector) on more than 11 million Amazon and Yelp reviews, we test whether reviews hold gender bias and determine the nature of the bias. Fourth in order to examine what harm can be caused if reviews hold gender bias, we conduct a pretest and a study. Finally, we discuss why it is important for firms to know that customer reviews contain gender bias, and how such bias can get perpetuated through stereotypically biased recommendations, messages, and insensitive portrayal. Consequently, the firm can face ethical issues of propagating a bias.

\section*{\large Theoretical Background}

\subsection*{\textit{\normalsize The harmful outcomes of bias}}

Stereotypes have been defined as the cognitive component of biased attitudes that people hold towards a majority or a minority group. Stereotypic attitudes are an intrinsic part of the social heritage of cultures and countries and it is difficult for humans not to be aware of the existence of such beliefs (Devine 1989; Greenwald et al. 1998). Stereotypic beliefs become harmful (or biased) when a negative attitude towards a minority results in prejudicial treatment. For instance, research has demonstrated that identical resumes (keeping education, experience, and skills the same) can result in different job offers by merely changing the name on the resume to reflect a racial group (Bertrand and Mullainathan 2004). Similarly, recent work has demonstrated that women are a) shown lower paying jobs than men while searching for the same type of jobs, b) less likely to be offered software jobs, and c) provided lower evaluations for the same work (Bohren et al. 2018; Bolukbasi et al. 2016; Datta et al. 2015). In the marketplace, biased behavior is visible when certain racial groups are perceived to have a higher likelihood of committing a crime and they are profiled in a discriminatory manner (Angwin et al. 2016). Similarly, research has documented the consequences of stereotypic biases on prisoner incarceration (longer incarcerations for African Americans having committed the same crime as Caucasians), parole decisions (longer paroles for minorities), loan approval (loan is less likely to be approved or approved for a smaller amount because of biases against certain racial or ethnic groups), etc. (Angwin et al. 2016; Miller 2015; Wang et al. 2013).

Two types of stereotypes have been considered in past research: explicit and implicit (Devine 1989; Greenwald et al. 1998). The former is an attitude that the individual is aware of holding, while the latter is when individuals may feel that they do not (explicitly) have any stereotypical bias about a group but do hold such a bias without being aware of it. Such a lack of awareness can cause the automatic activation of stereotype and make people complicit in the propagation of biased beliefs and behaviors in fields ranging from employment decisions to emergency medical treatment (Devine 1989; Devine et al. 2012; Green et al. 2007). It is a lot harder to detect implicit stereotypes because individuals themselves do not believe that they hold a biased attitude towards a group. However, it is important to detect them because lack of awareness means that individuals cannot consciously control for such biases creeping into their thoughts and actions, making biases more dangerous. For instance in recent research, science faculty felt that they were objectively assessing applicants for a lab manager position but implicit bias caused them to favor male applicants (Moss-Racusin et al. 2012). Implicit stereotypes, undetected, can cause the stereotype to propagate because the individual is not aware of holding the stereotype, though still influencing his decisions. Therefore, measures such as the IAT based on response time measures, rather than explicit measures have been designed to detect implicit stereotypical biases (Greenwald et al. 1998).

Applied to our context, we posit that individuals writing customer reviews may be implicitly transferring stereotypic biases to their reviews. Humans do not easily detect such biases because it is not explicit. However, as we discuss next, algorithms have the ability to detect signals from text that may not be explicit.

\subsection*{\textit{\normalsize Embedding of stereotypes in language}}

Language provides a window into people’s mind. The words people use when they speak or write has been used to detect their personality traits, their level of emotion, their confidence in the situation, and their honesty (Jurafsky, Ranganath, and McFarland 2009; Mairesse et al. 2007). It has been shown that extroverts tend to use more words, repetitions, positive emotion words, and informal language compared to introverts who use a broader vocabulary. Openness-to-experience individuals have been shown to use tentative words such as “maybe” or “might”. Depressed or stressed speakers use more first person singular pronouns (Rude, Gortner, and Pennebaker 2004).

Linguistics research suggests that words and the context in which they occur are embedded in cultural stereotypes (Liu and Karahanna 2017; Stubbs 1996). For instance, there are specific associations of women versus men with different attributes such as power, objectification, career associations, etc. (Holmes and Meyerhoff 2008). Many associations reliably appear in the language used. Using language from a speed-dating task, researchers were able to categorize individuals as flirtatious or friendly. The success of the categorization hinged on the fact that men and women use different words. Flirting women used more swear words while flirting men used more sexual vocabulary. Men labeled as friendly used “you” while women labeled as friendly used “I” (Jurafsky et al. 2009). Since stereotyping is a form of categorization, this research shows that language reflects the existing social stereotypes. Even if people do not intend to explicitly state a biased opinion, their words could implicitly reflect stereotypical biases.

Language captures one’s implicit stereotypic bias; hence, the bias insidiously makes its way into the reviews. Algorithms, which are commonly used to mine customer-generated content, have the ability to detect implicit bias expressed as opinions and sentiments (Domingos 2015). Algorithms can detect meanings, rules, signal hidden amid noise, and extract information about variables. Even if certain variables such as gender are removed from the dataset to ensure that it does not bias the results, an algorithm can detect them using other variables as proxy. For example, algorithms can detect sexual orientation from face expressions more accurately than humans (Wang and Kosinski 2017), and predict crime before it occurs (Wang et al. 2013). Therefore, if customer reviews contain implicit stereotypic bias, the algorithm can detect it. It becomes concerning because the algorithm considers any bias it has detected to be just a human preference and converts it into a rule (explicit) to be used in its decisions and recommendations. The bias becomes explicit and gets propagated.

\subsection*{\textit{\normalsize Finding gender stereotypes in customer reviews}}

Past research has used many different techniques to mine valuable information from customer-generated text. From traditional methods such as focus groups and surveys that specifically ask customers their thoughts about products to newer methods that unobtrusively and inexpensively “listen” to customer opinions by mining customer generated content. For instance, questions and answers posted by customers about vehicles have been used to glean their needs using a virtual Bayesian advisor (Urban and Hauser 2004). Latent Dirichlet Allocation has been used to mine consumer chatter (Tirunillai and Tellis 2014). Associative and clustering methods have been suggested for mining sentiments, opinions, and brand meanings from customer-generated content (Lee and Bradlow 2011; Netzer et. al. 2012). However, many of these clustering and semantic analysis methods do not extract meaning from user-generated text.

Recent research has demonstrated that word embedding methods are well suited for tasks that need to extract meaning from the context in which a word occurs (Mikolov et al. 2013; Pennington, Socher, and Manning 2014). A simple example helps explain word embedding algorithms. Humans and algorithms can easily discern that the colors red and maroon are more similar than the colors red and yellow. In contrast with humans, however, algorithms can detect the proximity of the colors red and maroon by calculating their distance on red, green, and blue dimensions. However, when trying to determine whether the word “rose” is closer to “pleasant” or “repulsive,” humans excel but algorithms struggle because words do not come with numerical dimensions attached to them. While the color red will be red in any context, the word “rose” may mean different things in different contexts. Word embedding algorithms solves this problem by assigning numerical dimensions to words to capture similarity among words.

An intuitive understanding of word embedding algorithms comes from Firth’s (1957: 11) hypothesis, “You shall know a word by the company it keeps.” That is, words that occur in similar contexts tend to have similar meanings. For example, if asked to think of a word that belongs with cow, drink, liquid, and calcium, most people would answer milk. Word embedding algorithms exploit the properties of co-occurring words to quantify semantic connections among them (i.e., by assigning semantically connected words near each other). This process results in each word being mapped onto a semantically relevant high-dimensional space and similar words appearing near each other. Word embeddings have been used extensively to document retrieval, entity recognition, understanding analogies, spam filtering, web search, trend monitoring, etc. (Mikolov et al. 2013; Pennington et al. 2014). In the current research, we use word embedding algorithm to find out whether female (as opposed to male) representing words appear more in the context of positive attribute or negative attribute words in customer reviews. 

We next describe the procedure and the results of our analysis.

\section*{\large Procedure}

\subsection*{\textit{\normalsize Data}}

We use two corpora to test whether algorithms designed to learn from text show gender bias: (1) 7.9 million reviews on movie DVDs posted on Amazon.com over a 15-year period and (2) 3.4 million Yelp reviews about restaurants. We use movie and restaurant reviews because people belonging to different groups and holding different identities universally consume both movies and food. In addition, both involve the presence of other people (i.e., movies have characters, and food is made by chefs and served by the wait staff).

\subsection*{\textit{\normalsize Using Global Vectors (GloVe) for Identifying Gender Bias}}

Word Embedding. In this research, we utilize GloVe (Pennington et al. 2014), a word embedding algorithm widely used to represent words as vectors. The vector representation captures the meaning of the word because it mathematically converts Firth’s (1957: 11) hypothesis, “You shall know a word by the company it keeps.” GloVe uses the context in which a word appears to represent a word as a point in a vector space. For example, we can represent the word Woman as a vector with 200 dimensions. If a word can be represented as a vector then the similarity between two words can be easily calculated using cosine distance. Stated briefly, GloVe would represent every unique word from the 7.9 million Amazon reviews and 3.4 million Yelp reviews as a vector with 200 dimensions. That is, each word would be converted into a vector uniquely identified by 200 numerical values. The numerical values on these 200 dimensions will capture the relationship of the occurrence of the words in the reviews. So if rose is considered more pleasant than unpleasant, we should find that the similarity of the word rose and pleasant would be higher than the word rose and unpleasant.

Appendix A lists technical details of GloVe and the advantages of using GloVe over other word embedding methods. It also shows how similarity is measured between words using cosine distance and how we used GloVe in this work.

Similarity and Bias. To determine whether the algorithm is picking up stereotypical biases from text corpora, we adapt a method suggested by Caliskan, Bryson, and Narayanan  (2017) that is modeled after the Implicit Association Test (IAT) (Greenwald et al. 1998; Greenwald, Nosek, and Banaji 2003). In IAT, if a target word is more strongly associated with a positive than a negative attribute in a person’s memory, he or she will respond faster to the pairing of the target word with a positive than a negative attribute.

Employing GloVe, a word is converted into a vector. In our case, the word Woman would appear as a unique vector with coordinates representing it in each of the 200 dimensions. Similarly, the word Man would be a represented as a unique 200-dimension vector, as would a positive attribute word Loyal and a negative attribute word Fickle. Cosine distance between any two vectors can be easily calculated and it would indicate the level of similarity between the two words such that 1 means the words are identical in meaning, 0 means they are not at all related (orthogonal) and -1 would mean they are opposite in meaning. Therefore, we use cosine similarity to find out whether male and female target words are more similar to positive versus negative attribute words. We use similarity between words in the same way as response time is used in IAT to test for gender bias.

Prior work on IAT has compared male and female names with career versus home/family-related attributes (Nosek, Banaji, and Greenwald 2002); we do the same. Let’s assume that there are two sets of male (e.g., John, Paul, Mike) and female (e.g., Amy, Joan, Lisa) target names. These target names are compared with two sets of attributes: one representing career (e.g., management, professional, corporation) and the other family (e.g., children, family, marriage). If an algorithm is picking up stereotypical biases in the given text corpus, male names would display greater similarity to career- than family-related attributes compared to female names. In this research, we also find the similarity of words related to men (e.g. he, him, son) or women (she, her, daughter) with negative (e.g. fickle, impulsive, selfish) and positive adjective (e.g. loyal, industrious, creative) attributes. Formally, we can represent the presence of such stereotypical biases as a net similarity measure.

\begin{center}
Similarity (male words to positive attributes) – Similarity (male words to negative attributes) $>$ Similarity (female words to positive attributes) – Similarity (female words to negative attributes) …    (1)
\end{center}

Such a representation enables us not only to state that a bias exists but also to examine the cause of the bias for that text corpus. For instance, did the bias occur because male names are equally similar to positive versus negative attributes but female names are more similar to negative attributes? Or are male names more similar to positive rather than negative attributes but female names are more similar to negative rather than positive attributes? We use the permutation test, which assigns p-value to each result, to examine whether obtained results are statistically significant. Details on the measurement of similarity, effect size calculation, and the permutation test are provided in Appendix B.

In both data sets, we perform gender bias tests. We conduct separate analyses for the Amazon.com and Yelp datasets. Again, the Amazon reviews were for DVD sales and contained 7.9 million reviews, while the Yelp reviews were for restaurants and contained 3.4 million reviews. We preprocess the data to remove punctuation and numbers and convert all words to lowercase. For each word, we create word vectors using GloVe such that each word is semantically represented as a 200-dimension vector. Next, we estimate similarities among words. To examine gender bias, we find similarities of target words (names or words signifying gender) to attribute words (“positive/negative” or “career/family”). We use the target and attribute words from prior work (Garg et al. 2018; Greenwald et al. 1998, 2003; Nosek et al. 2002; Weyant 2005) (see Appendix C for all the target and attribute words that we use).

\subsection*{\textit{\normalsize Validation}}

How do we know that the word embeddings in the Amazon or Yelp dataset correctly captures meaning? On each dataset, we perform a validation procedure suggested by past research (Caliskan et al. 2017; Greenwald et al. 1998, 2003; Nosek et al. 2002). The procedure suggests that if word embeddings are indeed capturing meaning, then known positive target stimuli (e.g., flowers) should display more similarity with positive attributes (e.g., pleasure, gentle) than negative attributes (e.g., grief, filth). But, known negative target stimuli (e.g., bugs) should display more similarity with negative than positive attributes. Referring to our prior description on net similarity, if people associate flowers with positive attributes and bugs with negative attributes then we should find that net similarity of flowers is higher than net similarity of bugs. Here, net similarity for flowers captures the difference between the similarity of flowers to positive attributes and flowers to negative attributes.

In the Amazon.com reviews, the results show that the association of positive versus negative attributes with flowers and bugs differ significantly (effect size d = 0.583, p $<$ 0.03). We next consider the nature of the difference using net similarity, which we calculate as the difference in association of flowers (or bugs) with positive compared to negative attributes. Higher value of net similarity indicates greater association with positive attributes. In this case, we find that the net similarity of flowers to positive attributes (0.046) is higher than net similarity of bugs to positive attributes (-0.061), (d = 0.107, p $\approx$ 0). However, the net similarity of bugs to negative attributes (0.07) to family is more than the net similarity of flowers to negative attributes (-0.223), (d = 0.312, p $\approx$ 0). Similar to Amazon.com reviews, in the Yelp reviews, the results show a bias against bugs (effect size d = 0.645, p $<$ 0.03). We find that the net similarity of flowers to positive attributes (-0.348) is higher than net similarity of bugs to positive attributes (-0.37), (d = 0.058, p $\approx$ 0). However, the net similarity of bugs to negative attributes (0.574) to family is more than the net similarity of flowers to negative attributes (-0.095), (d = 0.875, p $\approx$ 0).
 These results help us validate our method and confirm that the word embedding algorithm is able to reliably extract meanings from both Amazon and Yelp reviews.

\subsection*{\textit{\normalsize Results: Amazon.com reviews}}

We next test for gender bias in four comparisons where we compare the similarity of male/female names (e.g. John, Paul versus Joan, Lisa), and of male/female words (e.g. he, him versus she, her) with both family and career words (e.g. management, professional versus family, children) and as well as positive and negative attributes (e.g., rational, innovative versus impulsive, conformist). Please see Appendix C for the complete list of words.

Male/female names with career/family words. When we consider the similarity of male versus female names to career- versus family-related attributes, we find a gender bias (effect size d = 1.382, p $<$ .001). As mentioned previously, examining the cause of the stereotypical bias is important because bias can be driven by bias against or preference for a target group. In this case, we find that gender bias occurs because the net similarity of male names to career (1.244) is higher than net similarity of female names with career (0.694), (d = 1.467, p $\approx$ 0). However, the net similarity of female names (1.617) to family is more than the net similarity of male names to family (1.508), (d = 0.384, p $\approx$ 0). Therefore the results indicate that gender bias occurs because women are less associated with career-related attributes and more with family-related attributes compared to men. If these reviews were used to provide recommendations in the marketplace, it would result in women being recommended less career-oriented products e.g. less online courses, job advertisements, or even less paying jobs (e.g., Bolukbasi et al. 2016).

Male/female names with positive/negative attributes. Next when we consider the similarity of male versus female names to positive and negative attributes, we again find a gender bias (d = 1.688, p $\approx$ 0). Specifically, when we consider what drives the gender bias we find: male names are more associated with positive attributes (0.692) than female names (0.344), (d = 1.413, p $\approx$ 0). However, female names have a higher net similarity with negative attributes (0.119) than male names (-0.075), (d = 1.402, p $\approx$ 0). Again, we find that gender bias occurs because of a positive bias towards men as well as a negative bias against women.

Male/female words with career/family words. We examine the similarity of male versus female words (e.g. he, him versus she, her) to career and family related words; we again find a gender bias (d = 0.828, p $<$ .03). While examining the cause of the stereotypical bias, we find that gender bias occurs because of higher net similarity of male words with career (2.227) compared to female words with career (1.475), (d = 0.862, p $\approx$ 0). Unlike in previous instances, in this case we find that male words have a higher net similarity to family words (3.009) compared to female words (2.902), (d = .121, p $\approx$ 0). The gender bias still emerges because the difference in net similarity of men with career outweighs the net similarity of men with family. However, this is less so for women.

Male/female words with positive/negative attributes. Our final comparison is the similarity of male versus female words to positive and negative attributes; we again find a gender bias (d = 1.012, p $<$ .01). When we consider what drives the gender bias we find that male names have a higher net similarity with positive attributes (1.707) than female words (1.241), (d = 0.904, p $\approx$ 0). However, female words are more similar to negative (0.769) rather than male words (0.746) attributes (d = 0.086, p $\approx$ 0).  Again, we find that the bias is driven because of higher association of men to positive attributes and women to negative attributes.

\subsection*{\textit{\normalsize Results: Yelp reviews}}

Male/female names with career/family words. When we consider the similarity of male versus female names to career- versus family-related attributes, we find a gender bias (d = 1.424, p $<$ .001). In this case, we find that gender bias occurs because the net similarity of male names to career (1.241) is higher than net similarity of female names with career (0.574), (d = 1.454, p $\approx$ 0). The net similarity of male names to family (0.801) is also more than the net similarity of male names to family (0.499), (d = 1.042, p $\approx$ 0). However, the difference in the similarity of male and female names to career is far more than the difference in similarity with family words.

Male/female names with positive/negative attributes. Next when we consider the similarity of male versus female names to positive and negative attributes, we again find a gender bias (d = 1.559, p $\approx$ 0). Specifically, when we consider what drives the gender bias we find: male names are more associated with positive attributes (0.619) than female names (0.147), (d = 1.465, p $\approx$ 0). However, female names have a higher net similarity with negative attributes (-0.103) than male names (-0.459), (d = 1.508, p $\approx$ 0). Again, we find that gender bias occurs because of a positive bias towards men as well as a negative bias against women.

Male/female words with career/family words. We examine the similarity of male versus female words to career and family related words. However, in this case we do not find a statistically significant gender bias (d = 0.301, p = 0.25).

Male/female words with positive/negative attributes. Our final comparison is the similarity of male versus female words to positive and negative attributes; we find a gender bias (d = 0.822, p $<$ .033). When we consider what drives the gender bias we find that male names have a higher net similarity with positive attributes (1.045) than female words (0.839), (d = 0.464, p $\approx$ 0). However, female words are more similar to negative (-0.059) rather than male words (-0.356) attributes (d = 1.111, p $\approx$ 0).  Again, we find that the bias is driven because of higher association of men to positive attributes and women to negative attributes.

In sum, across both Amazon and Yelp reviews, we find evidence of gender bias. Overall gender bias emerges, because of higher similarity of men with positive and career related attributes than negative and home-related attributes. For women the similarity to negative attributes or family-related words is higher. Therefore, the results indicate that the gender bias would result in men being associated more with positive attributes and less with negative attributes; however, women would be less associated with positive and more with negative attributes. 

\section*{\large So what?}

So what if reviews are biased against women? Even if the algorithm is learning biased preferences against women, how do we know that there are any consequences of this learned bias? The vast majority of algorithms used by Google, Amazon etc. are not accessible to the public and it is virtually impossible to peer inside them for any bias. However, it is possible to look at their output in response to different search queries and assess if bias against women shoppers has made an impact. Following past work (Angwin et al. 2016), we adopt a procedure to assess bias in commercially implemented algorithms.

If it is true that algorithms are learning to associate women with negative marketing attributes compared to men, then we should see this association appear even in non-recommendation settings. We use search queries with negative attributes (e.g., impulsive, conformist) vs. positive attributes (e.g., sensible, innovative) on Google to see whether the images that are shown by Google have more depictions of women or men. We count the number of times women are shown versus men are shown on the first screen. The image-search results across impulsive shopper, conformist shopper, lazy shopper, emotional shopper, and vain shopper are aggregated under negative attribute queries. While those in sensible shopper, innovative shopper, industrious shopper, rational shopper, and determined shopper are aggregated under positive attribute queries. Images that did not have men or women are classified as “other”. The null hypothesis is that there is no relationship between positive and negative attributes search queries and male and female images shown.

The results reject the null hypothesis ($\chi^{2}$ = 46.59, p $<$ .0001). To examine further, we estimate standardized residuals following the procedure suggested by Agresti (2007). We obtain large positive residuals for negative-attribute-queries depicting females images (6.81) and positive-attribute-queries depicting male images (1.19). This shows that more female images appear when searching for negative attributes (20\%) than positive attributes (11.6\%). However, more male images appear when searching for positive attributes (6.68\%) than negative attributes (5.71\%). A similar pattern of results appear when we partition the data table and compare results of positive and negative attributes for only male and female images (i.e., when we do not consider “other” images). The Log likelihood ratio test (G-test with 1 df = 18.83, p $<$ .001) provides the same conclusion.

One could argue that while gender bias exists and manifests in algorithm output, there is no reason to worry that such bias will have any influence on actual consumer choice. The rationale for this argument comes from past work that shows that people don’t rely on algorithmic results. Instead they might avoid suggestions/recommendations made by algorithms (Dietvorst, Simmons, and Massey 2015). However, it is worth noting that the results are mixed on algorithmic recommendation-acceptance given recent work which has shown that people do accept algorithmic advice (Logg, Minson, and Moore 2018). In response to these mixed findings, we adopt a procedure that is generally used to create recommendation systems. Then we test the influence of biased recommendations on consumer choice. We discuss it next.

Algorithms make recommendations through a matching process e.g. if Alex likes action movies and Jurassic World has a label attached to it that says it is an action movie, then through a simple match, the algorithm would recommend this movie to Alex. On sites like Netflix, any movie title has more than 76,000 labels attached to it (Madrigal 2014). However, products on their own do not have any labels. Attaching labels to a product like calling Jurassic World an action movie is done by humans. In fact, we now know that human assigned labels can significantly enhance quality of algorithmic product recommendations (Chang et al. 2016).

To assign labels such as action movie, or that diapers’ potential buyers are women, one of the procedures is to crowdsource consumer knowledge to generate labels to improve product recommendations (Costa et al. 2011; Harper et al. 2015; Heer and Bostock 2010). Our analysis till now has provided us with attributes associated to women versus men. Therefore, we asked people on the crowdsourced website MTurk to list the products they associate with a specific attribute. Specifically, in a pretest we asked people what products they associate with positive and what with negative attributes, and then used their recommendations as labels in a subsequent recommendation study. The aim of the recommendation study was to test whether algorithmic product recommendations have any influence or not.

\vspace{6mm}
 \section*{\large Pretest}

One challenge that we have is that the set of potential products is very large (e.g., Amazon alone sells hundreds of thousands of different products) making it difficult to determine, what products to use in our recommendation study. To address this challenge, we relied on the fact that we can create a finite set of attributes. We showed sixty participants either positive or negative attributes and asked them to list 2 products that they would recommend for a consumer described by such attributes. For instance, we asked to recommend two products for an “impulsive consumer” or two products for a “sensible consumer”. The positive attributes we used were innovative, determined, sensible, rational, certain, and industrious while the negative attributes were conformist, vain, impulsive, emotional, risky, and lazy. Many recommendations were nonspecific such as books or clothing, inappropriate such as guns, and nonsensical such as chicken. Many were about the same product but described differently such as discounted chocolates, inexpensive chocolates. We combined the labels that had commonality to find the most recommended product. Based on the recommendations provided by the participants of the pretest, we constructed the stimulus for the main recommendation study as indicated in the table below.

\begin{center}
  \begin{tabular}{ |p{2cm}|p{4cm}|p{2cm}|p{4cm}|  }
 \hline
 Positive Attribute & Product & Negative Attribute & Product\\
 \hline
 Sensible   & Discounted store Chocolate & Impulsive & Pack of Godiva Chocolate\\
  \hline
  Certain & A bond fund that offers guaranteed 3\% return per year & Risky & A stock fund that offers 6\% return with 50\% chance of getting 6\% and 50\% chance of getting 0\%\\
   \hline
 Rational & Telescope & Emotional & Digital photo frame\\
  \hline
 Innovative & Easel & Conformist & Business pens\\
  \hline
 Determined & Language course & Vain & Selfie stick\\
  \hline
 Industrious & Helpful tool box & Lazy & Luxurious Pillow\\
\hline
\end{tabular}
\end{center}

\vspace{4mm}
\section*{\large Recommendation Study}

As depicted in the table above, we created dyads of products such that one product mapped on to a positive adjective and the other to a negative adjective. Our word embedding analysis has demonstrated that negative attributes are more likely to be associated with female consumers than male consumers and vice versa. If this were converted into product recommendations, then algorithms are more likely to recommend products associated with negative attributes to female consumers compared to male consumers. The aim of this study is to demonstrate the outcome of such a recommendation. If consumers are more likely to accept a product recommendation, it demonstrates that the gender bias learnt by algorithms through moderated customer reviews is propagated and influence consumer choice. However, if people are averse to following the advice of algorithms, then we can conclude that it doesn’t matter even if algorithms are detecting gender bias in the reviews. 

\vspace{5mm}
\subsection*{\textit{\normalsize Procedure}}
We recruited three hundred and forty one participants from M-Turk and from a university participant pool and randomly assigned them to three conditions: control condition with no recommendation, positive-algorithmic recommendation, or negative-algorithmic recommendation. Those in the recommendation condition (positive as well as negative) first provided their demographic (gender) information. They were informed that their response to this question would help us understand their preference so that we can recommend products to them (similar to Netflix asking movie genre preference from a new customer). Such a process makes our test more conservative because consumers would know that their gender information is being factored into the recommendation. They can be more watchful of the recommended option and reject it if they feel that it is biased.

Subsequently, we informed participants that the algorithm has learnt their unique preference and will provide personalized recommendation to them for some product decision that consumers make regularly. Each product decision included a dyad of products that were priced the same but one was associated with a positive attribute while the other was associated with a negative attribute. The product dyads were as shown in table 1. The algorithm-recommended option was indicated with an arrow. Participants were free to accept the recommended product or choose the other product. In the positive-recommendation condition, an arrow indicating the recommendation appeared with the products associated with positive attributes. However, in the negative-recommendation condition, the arrow appeared with the products associated with negative attributes. The control condition asked consumers to choose between the product dyads but they were neither informed that the products were chosen to match their unique preference nor was any recommendation provided.

\subsection*{\textit{\normalsize Results}}

We ran a repeated measure logistic regression with condition (control, positive-recommendation, negative-recommendation) and gender (female, males) as independent variables. The results indicate a main effect of condition ($\chi^{2}$ = 27.41, p $<$ .0001), a main effect of gender ($\chi^{2}$ = 9.25, p $<$ .002) and an interaction between gender and condition ($\chi^{2}$ = 8.71, p $<$ .02). Next we decompose the interaction for men versus women.

Comparing the influence of recommendation among women, we find that their choice for the positive-attribute product in the positive-recommendation condition (61.81\%) is not significantly different from the control condition (58.65\%) (log odds ratio = -0.14, z ratio = -0.77, p = 0.72). However, their choice for the positive-attribute product significantly drops in the negative-recommendation condition (41.67\%) compared to the control (58.65\%) (log odds ratio = -0.71, z ratio = -4.04, p $<$ 0.001). This shows that while positive-recommendation does not increase the choice of positive-attribute product (e.g., discounted store chocolate or language course), negative recommendation decreases the choice of the positive-attribute product. Thus, negative recommendation influences women’s choice more than positive recommendation.

The pattern of results was different for men. When comparing the influence of recommendation among men, we find that men are more likely to choose the positive-attribute product in the positive-recommendation condition (63.84\%) than in the control condition (54.1\%) (log odds ratio = -0.42, z ratio = -2.48, p $<$ 0.04). However, in the negative-recommendation condition men were as likely to choose the positive-attribute product (54.16\%) as in the control condition (54.1\%), (log odds ratio = 0.002, z ratio = 0.16, p = 0.99). This shows that positive recommendation influences men’s choice more than negative recommendation.

The findings of this study demonstrate that gender bias in recommendations can critically impact consumer choice. 

\section*{\large Conclusion}

Research from diverse fields have demonstrated that algorithms are being used to make several life-altering decisions from fighting crime to job candidate selection to loan approvals (Jannach et al. 2010; Owen 2015; Schrage 2014). Therefore, it is important to know the nature of the data that these algorithms are learning from. If the data is biased, recommendations and decisions will perpetuate the same gender bias. In this research, we find that algorithms infer gender bias in moderated customer reviews. Our research findings add to findings across disciplines that show that algorithms are predicting outcomes that are biased against certain groups (Angwin et al. 2016; Bolukbasi et al. 2016; Schrage 2014). Moreover, we answer the “so what” if reviews are biased with two pieces of evidence. First, we show that more female images are shown when the search-query is negative (e.g. impulsive shopper) than positive (e.g. sensible shopper). Second, we demonstrate that in the domain of product recommendation, biased recommendations can influence choices made by female consumers. Among female consumers, while positive-recommendation does not increase the choice of positive-attribute product, negative recommendation decreases the choice of the positive-attribute product. Thus, negative recommendation influences women’s choice more than positive recommendation. Conversely, positive recommendation influences men’s choice more than negative recommendation.

Our findings have ethical and legal ramifications for firms. Firms routinely gather insights from customer reviews assuming them to be the voice-of-the-customer but if these reviews hold gender bias, then any decision made using those reviews has the power to harm customers, especially female customers. There has been much focus on ethicality of algorithmic decisions given the use of a black-box approach that does not reveal all paths to the final decision (Angwin et al. 2016; Bostrom and Yudkowsky 2011). For instance, if a firm uses reviews containing gender bias to design products and promotions, recommend products to consumers, determine how to portray individuals in a commercial, segment consumers etc. then it could lead to biased recommendations, messages or insensitive portrayals. Specifically, recommendation using biased reviews would result in learning and using consumer vulnerability against female consumers. Moreover, if these reviews are used to provide recommendations in the marketplace, it would result in women being recommended less career-oriented products e.g. less online courses, job advertisements, or even less paying jobs (e.g., Bolukbasi et al. 2016).

Illuminating algorithmic biases has implications beyond the contexts discussed in this research. A massive amount of data on citizens is collected on a daily basis. For example, the New York Police Department has more than 6000 surveillance cameras and license plate readers (Owen 2015); private firms are deploying hundreds of satellites to take pictures of every part of the earth to create real-time surveillance tools. With perpetual surveillance of everyday life by government and non-government entities, a phase has been reached in which enough data on consumers have accumulated to create a 360-degree view of them. If algorithms are going to process these data to make decisions that inform government agencies, it is easy to imagine similar biases emerging in policy and law enforcement decisions.

If algorithms are influencing so many aspects of decision-making, they should not come with the risk of caveat emptor. The first step in fighting stereotypic bias is to become aware that bias exists and be concerned about the harm that such a bias can cause (Devine et al. 2012). Once these two conditions are met, interventions have a chance to be successful. Applied to our context, we suggest that becoming aware that reviews contain stereotypic bias and subsequent harm to consumers and firms is important.

A limitation of our research is that we have considered only English language text; it would be worthwhile to examine whether reviews in other languages might also hold (or not) such gender bias.

\newpage
\begin{center}\section*{\large Appendix A: Description of Global Vectors GloVe}\end{center}

We provide a brief overview of GloVe, a word embedding algorithm (Pennington et al. 2014). The primary aim of word embedding is to provide numerical representation to each word so that the information contained in human language can be analyzed like any other numerical data. 
Why do we need to have a unique numerical representation of a word? We need this for two reasons: 1) converting a word to a unique numerical representation helps us analyze words just like numbers e.g. finding similarity via distance and 2) if the unique representation of a word captures the meaning of the word, then this allows us to move beyond a count-based approach (which considers each word to be a presence versus absence and nothing about its meaning) to compare two words based on their meaning. This quality of word embedding distinguishes it from other text analysis methods like topic modeling (Blei, Ng, and Jordan 2003\footnote{Blei, David M., Andrew Y. Ng, and Michael I. Jordan. Latent dirichlet allocation. \textit{Journal of machine Learning research 3}, no. Jan (2003): 993-1022.}) that can find out latent topics in a document or words in a topic but they are unable to compare words, based on their meanings, to each other. Topic models cannot be used if we want to know the semantic (meaning-based) similarity of two words e.g. is cake considered more tasty than a salad; is Apple considered more premium than Dell and in our case, are female consumers considered more impulsive than male consumers? For such tasks where we want to obtain semantic associations, relations or similarity between two words, word embeddings are used.

In essence, in embedding models, words are assigned a position in a vector space based on the context that a word shares with other words in the corpus. Words that share many contexts are positioned near one another in this space (e.g., rose and pleasant), while words that inhabit very different contexts are located farther apart (e.g., rat and pleasant). Words frequently sharing linguistic contexts, and thus located nearby in the vector space, tend to share similar meanings. Co-occurrence is considered an indicator of meaning.

\vspace{5mm}
\subsection*{\textit{\normalsize How does GloVe work?}}

GloVe first creates a co-occurrence matrix by counting the co-occurrence of each word with every other word in the corpus within a context window. The context window is defined as 10 words before or after the target word. For instance, if our document has two pangrams, “The quick brown fox jumps over the lazy dog. A mad boxer shot a quick, gloved jab to the jaw of his dizzy opponent”. Then a context window of 10 for the word “mad” will have 10 words that precede it “The quick brown fox jumps over the lazy dog” and 10 words that succeed it “boxer shot a quick, gloved jab to the jaw of”. For each word, the context window would contain different words. Context windows help move beyond a simple count based approach in text analysis and bring in Firth’s hypothesis that the meaning of a word can be captured by the words that appears along with it.

We explain GloVe using an example relevant to the manuscript. In order to test for gender bias, we examine whether female (rather than male) words occur more in the context of negative rather than positive attributes in a corpus. We first combine all the reviews, say for Amazon, and create one document. Then we create a matrix that counts the number of times a target word e.g. she (female word) or he (male word) occurs in the context of fickle (negative attribute) or loyal (positive attribute) in the document. If we define a context window of 10 then 10 words before and 10 words after would be considered for counting co-occurrence. GloVe moves beyond just a count and calculates the probability $P_{ji}$ of a target word j occurring in the context of the attribute word i by using P(j$|$i) = $\frac{X_{ji}}{X_{i}}$. Here $X_{ji}$ is the number of times word j appears the context of word i and $X_{i}$ is the number of times any word in the document appears in the context of i.

If gender bias exists in the reviews, then we would find that the probability of \textit{she} (denoted by j) occurring in the context of \textit{fickle} (denoted by i) is more than the probability of \textit{he} (denoted by k) occurring in the context of \textit{fickle} $P_{ji}$  $>$ $P_{ki}$ or $\frac{P_{ji}}{P_{ki}}$ $>$ 1. Or we would find the probability of \textit{he} occurring in the context of \textit{loyal} (denoted by l) more than the probability of \textit{she} occurring in the context of \text{loyal} $P_{kl}$  $>$ $P_{jl}$ or $\frac{P_{kl}}{P_{jl}}$ $>$ 1. However, for words unrelated to both men and women e.g. tree (denoted by t) the probability of occurring is likely to be similar in the context of both \textit{fickle} and \text{loyal} $P_{tl}$  = $P_{ti}$ or $\frac{P_{tl}}{P_{ti}}$ = 1. The use of ratio of probability co-occurrences rather than simple co-occurrences count helps discriminate relevant from irrelevant words and thus, helps capture the meaning between two words.

The ratio of probability co-occurrences depends on three words i, j, and k and hence the general model can be represented as F ($w_{i}$, $w_{j}$, $\overrightarrow{{w}_{k}}$) =  $\frac{P_{ik}}{P_{jk}}$. Here, $w_{i}$ is the word vector for word i and $w_{j}$ is the word vector for word j. This general model develops into a weighted least squares cost function. Using stochastic gradient descent method, minimizing the cost function yields vectors (numerical dimensions) for each word. Words can be represented as vectors with any number of dimensions but a smaller number of dimensions like 30 or 50 lose information. Thus, in this work we convert words to 200 dimensions. 

\subsection*{\textit{\normalsize Using word vectors to measure similarity}}

Once we have converted a word into a vector, we can now mathematically use it in our analysis. The simplest thing we can do is to find out the similarity between two words. Just as in any unsupervised learning method, the distance between words is used to measure similarity. However, unlike in 2 or 3 dimensions when we use Euclidean distance, in this high-dimensional space we use the cosine of the angle between word vectors to measure similarity.

{Cosine Similarity = $\frac{\sum_{i=1}^{n} x_{i}y_{i}}{\sqrt{\sum_{i=1}^{n} x_{i}^{2}}{\sqrt{\sum_{i=1}^{n} y_{i}^{2}}}}$

Values of cos$\theta$ can vary from +1 to -1 because cos(0) = 1, cos(90) = 0 and cos(180) = -1. For words that are very similar, cosine distance will be near 1. As words become more dissimilar, cosine distance will decrease and become nearly 0 (moving from 1 to 0), whereas when words are opposite in meaning, it will become negative (moving from 0 to -1).

\newpage
\begin{center}\section*{\large Appendix B: Method to test for gender bias}\end{center}

In order to examine gender bias, we adapted a method that is modeled after the Implicit Association Test (IAT) but applied for words in text analysis (Caliskan et al. 2017). Let’s take a simple example of assessing the presence of gender bias to understand how this method works. Assume we have two sets of target words: male (e.g. he, him, boy) and female (e.g. she, her, girl). We then compare the target words to two sets of attribute that are negative (e.g. fickle, impulsive, lazy) versus positive (e.g. loyal, sensible, industrious). The intuitive idea behind this method is that if gender bias does exist, then we should find that female words display higher similarity to negative attributes compared to male words (or male words display higher similarity to positive attributes compared to female words).

Formally, let X represent a set of male words and Y a set of female words. Let A be the set containing positive attributes and B negative attributes. We use cosine distance to measure similarity between two word vectors. Therefore, for a male word x (member of set X) and an attribute word a (member of set A containing positive attribute) $\cos (x,a)$ will give us the value of similarity between x and a, where $\cos(x,a)$ =  $\frac{x\cdot a}{\mid x \mid\cdot \mid a \mid}$ and $x\cdot a$ is dot product of vectors x and a. Hence, we can calculate the similarity between x and each of the members of set A (i.e., positive attribute). We can then calculate the average similarity between the target word x and all the positive attributes in set A. $Mean_{a\epsilon A}{\cdot\cos(x,a)}$ denotes this average similarity to all the positive attributes. Similarly, the estimated average similarity of x to all the negative attributes in set B would be given by $mean_{b\epsilon B}{\cdot\cos(x,b)}$. 

A difference of $mean_{a\epsilon A}{\cdot\cos(x,a)}$ and $mean_{b\epsilon B}{\cdot\cos(x,b)}$ would provide a single measure of relative similarity of word x to positive versus negative attributes. Intuitively, such a difference will provide net similarity to positive attributes. If $mean_{a\epsilon A}{\cdot\cos(x,a)}$ $-$ $mean_{b\epsilon B}{\cdot\cos(x,b)}$ $>$0 then it would show that x is closer to positive rather than negative attributes. Conversely, if $mean_{a\epsilon A}{\cdot\cos(x,a)}$ $-$ $mean_{b\epsilon B}{\cdot\cos(x,b)}$ $<$0  then it would show that x is closer to negative rather than positive attributes. If S(x,A,B) = $mean_{a\epsilon A}{\cdot\cos(x,a)}$ $-$ $mean_{b\epsilon B}{\cdot\cos(x,b)}$ then we can calculate S(x,A,B) for all the members of the set X. For set X, $\sum_{x\epsilon X}{\cdot S(x,A,B)}$ captures the sum of this net similarity to positive attributes for all of its members. 

Similarly, the net similarity of all the members of set Y (i.e. set of female words) to positive and negative attributes is given by $\sum_{y\epsilon Y}$ S(y,A,B) . The main measure of a gender bias would then be S(X,Y,A,B)= $\sum_{x\epsilon X}$ S(x,A,B) $-$ $\sum_{y\epsilon Y}$ S(y,A,B). A positive value of S(X,Y,A,B) would show that names in set X (male words) are more similar to positive attributes than names in set Y (female words). However, a negative value would show that words in set Y are closer to positive attributes than those in set X. 

However, S(X,Y,A,B) is just one measure of relative similarity and one could argue that it is an outcome of a random process. Therefore, we need to rule out the null hypothesis that the similarity measure was obtained by a random partition of names in two sets. Stated differently, we need to estimate the probability that the obtained similarity has not emerged due to a random shuffling of names in set X and Y.  Such probability can be calculated by 1) obtaining similarity score for all the partitions of the given six names in two sets i.e., creating sets like X = {he, his, boy} and Y= {she, her, girl}, and 2) finding the number of times such partitions gave a score higher than the obtained score.

The process described above is essentially a non-parametric permutation test where if the null hypothesis is true, then randomly shuffling names in the two sets should lead to S(X,Y,A,B) scores that are no different than the obtained score.

Formally, if $(X_{i}, Y_{i})$ represents the potential random shuffling of names in set X and Y then probability of obtained score being an outcome of random process will be

\vspace{3mm}
\begin{center}{Probability = $\frac{Number of times (S(X_{i},Y_{i},A,B) > \mid{S(X,Y,A,B)\mid)}}{Number of all potential shuffling of set X and Y}$}\end{center}

This will give us a two-tailed p-value. It is important to note that the denominator of the probability estimate can get very large if we have many members in set X and Y. In such situations, a sufficiently high number of shuffles provide an approximate value of probability. For each comparison reported in this work, we did 5000 shuffles of names in two sets.

The effect size of the stereotypic bias can be estimated by

\begin{center}{$\frac{Mean_{x\epsilon X} S(x,A,B) - mean_{y\epsilon Y} S(y,A,B)}{StandardDeviation_{w\epsilon X\bigcup Y} S(w,A,B)}$}\end{center}

It is a normalized estimate of the stereotypic bias similar to Cohen’s \textit{d}. 
The following is the summary of the process used to examine if algorithms contain stereotypic biases.

\begin{enumerate}
\item We created separate word embedding for movie reviews and restaurant reviews of each of 200 dimensions.
\item We tested for gender biases that have been represented and tested using only words i.e. we did not test for gender bias measured via images.
\item The target words were names or descriptive words, whereas the attribute words were career/family based or positive/negative attribute associations. All words were adopted form past research.
\item We used the WEAT procedure suggested by Caliskan et al. (2017) to find how two sets of words, target and attributes are associated with each other meaningfully.
\end{enumerate}

\newpage
\begin{center}\section*{\large Appendix C: The names and target attributes used to test for gender bias}\end{center}

\vspace{3mm}
The target words and attribute words appear in lowercase as we used them in the analysis.

\subsection*{\textit{\small{Female/Male target names}}}
\textit{Female target names}: joan, lisa, sarah, diana, kate, ann, amy, donna

\textit{Male target names}: john, paul, mike, kevin, steve, greg, jeff, bill

\subsection*{\textit{\small{Career/Family words}}}
\textit{Profession attribute words}: executive, management, professional, corporation, salary, office, business, career

\textit{Family/home attribute words}: home, parents, children, family, cousins, marriage, wedding, relatives

\subsection*{\textit{\small{Female/Male target words}}}
\textit{Female target words}: she, hers, her, woman, female, herself, women, females, gal, girl 

\textit{Male target words}: he, his, him, man, male, himself, men, males, guy, boy

\subsection*{\textit{\small{Positive/Negative attribute words}}}
\textit{Positive attribute word}: Honest, reasonable, independent, thorough, dependable, rational, relaxed, loyal, reliable, disciplined, patience, creative, innovative, planned, resolute, resisted, industrious, creative, determined, wise, tough, jolly, civilized, strong, enterprising, quick, logical, original, methodical, kind
    
\textit{Negative attribute words}: unfriendly, unkind, rigid, moody, intolerant, hedonistic, tempted, fragile, indulgent, irresponsible, instinctive, dissatisfied, conformist, impulsive, fickle, unreliable, emotional, conformist, vain, lazy, submissive, irritable, frivolous, inhibited, sensitive, vindictive, complicated, changeable, sarcastic

\subsection*{\textit{\small{Validation words and attributes}}}
\textit{Flowers}: aster, clover, hyacinth, marigold, poppy, iris, orchid, rose, daffodil, lilac, pansy, tulip, buttercup, daisy, lily, peony, violet, carnation, magnolia, petunia, bluebell.
 
\textit{Insects}: caterpillar, flea, locust, bedbug, centipede, fly, maggot, tarantula, bee, cockroach, gnat, mosquito, termite, beetle, cricket, hornet, moth, dragonfly, roach.

\textit{Positive attribute words}: caress, freedom, love, peace, cheer, friend, heaven, loyal, pleasure, diamond, gentle, honest, lucky, rainbow, diploma, gift, honor, miracle, sunrise, happy, laughter, paradise, vacation.

\textit{Negative attribute words}: abuse, crash, filth, murder, sickness, accident, death, grief, poison, stink, assault, disaster, hatred, pollute, tragedy, bomb, divorce, jail, poverty, ugly, cancer, evil, kill, rotten, vomit.

\end{document}